\documentclass[conference]{IEEEtran}
\IEEEoverridecommandlockouts
\usepackage{cite}
\usepackage{amsmath,amssymb,amsfonts}
\usepackage{algorithmic}
\usepackage{graphicx}
\usepackage{textcomp}
\usepackage{xcolor}
\usepackage{multirow}
\usepackage[inline]{enumitem}
\def\BibTeX{{\rm B\kern-.05em{\sc i\kern-.025em b}\kern-.08em   T\kern-.1667em\lower.7ex\hbox{E}\kern-.125emX}}
\begin{document}

\title{
Multimodal Carotid Risk Stratification with Large Vision-Language Models: Benchmarking, Fine-Tuning, and Clinical Insights
}

\author{\IEEEauthorblockN{1\textsuperscript{st} Daphne Tsolissou}
\IEEEauthorblockA{\textit{National Technical University of Athens} \\
\textit{Archimedes, Athena Research Center}\\
Athens, Greece \\
dtsolisou@biosim.ntua.gr
}
\and
\IEEEauthorblockN{2\textsuperscript{nd} Theofanis Ganitidis}
\IEEEauthorblockA{\textit{National Technical University of Athens} \\
Athens, Greece \\
orcid.org/0009-0006-7794-9793}
\and
\IEEEauthorblockN{3\textsuperscript{rd} Konstantinos Mitsis}
\IEEEauthorblockA{\textit{National Technical University of Athens} \\
Athens, Greece \\
orcid.org/0000-0002-4629-2163}
\and
\IEEEauthorblockN{4\textsuperscript{th} Stergios Christodoulidis}
\IEEEauthorblockA{\textit{CentraleSupélec, Université Paris-Saclay} \\
Paris, France \\
stergios.christodoulidis@centralesupelec.fr}
\and
\IEEEauthorblockN{5\textsuperscript{th} Maria Vakalopoulou}
\IEEEauthorblockA{\textit{CentraleSupélec, Université Paris-Saclay} \\
\textit{Archimedes, Athena Research Center}\\
Paris, France \\
maria.vakalopoulou@centralesupelec.fr
}
\and
\IEEEauthorblockN{6\textsuperscript{th} Konstantina Nikita}
\IEEEauthorblockA{\textit{National Technical University of Athens} \\
Athens, Greece \\
orcid.org/0000-0002-4629-2163}
}
\maketitle


\begin{abstract}
Reliable risk assessment for carotid atheromatous disease remains a major clinical challenge, as it requires integrating diverse clinical and imaging information in a manner that is transparent and interpretable to clinicians. This study investigates the potential of state-of-the-art and recent large vision–language models (LVLMs) for multimodal carotid plaque assessment by integrating ultrasound imaging (USI) with structured clinical, demographic, laboratory, and protein biomarker data. A framework that simulates realistic diagnostic scenarios through interview-style question sequences is proposed, comparing a range of open-source LVLMs, including both general-purpose and medically tuned models. Zero-shot experiments reveal that even if they are very powerful, not all LVLMs can accurately identify imaging modality and anatomy, while all of them perform poorly in accurate risk classification. To address this limitation, LLaVa-NeXT-Vicuna is adapted to the ultrasound domain using low-rank adaptation (LoRA), resulting in substantial improvements in stroke risk stratification. The integration of multimodal tabular data in the form of text further enhances specificity and balanced accuracy, yielding competitive performance compared to prior convolutional neural network (CNN) baselines trained on the same dataset. Our findings highlight both the promise and limitations of LVLMs in ultrasound-based cardiovascular risk prediction, underscoring the importance of multimodal integration, model calibration, and domain adaptation for clinical translation.
\end{abstract}


\section{Introduction}
Recent advances in Artificial Intelligence (AI), have led to the development of Large Vision-Language Models (LVLMs), a class of multimodal systems designed to jointly process visual and textual information \cite{li2025vision}. Such models combine encoders, typically Vision Transformers (ViT) \cite{vit} specialized in image inputs, with Large Language Models (LLMs), to encode text inputs. This combination has shown remarkable capabilities in processing and integrating information across multiple modalities \cite{li2025vision}. In natural images, models such as LLaVa-NeXT \cite{llava-next} have been successfully applied to a wide range of tasks, including image captioning, visual question answering (VQA), report generation, and multimodal retrieval. These reasoning abilities emerge from a process called chain-of-thought (COT) prompting, a process where a model generates intermediate natural language reasoning steps before producing a final answer \cite{wei2023cot}. Interview-style prompting, in which the model is guided through step-by-step analysis, has been shown to elicit deeper modality-specific reasoning and improve performance in zero-shot settings \cite{xu2025llavacot}.

\begin{figure*}[htbp]
\centerline{\includegraphics[scale=0.4,keepaspectratio=true]{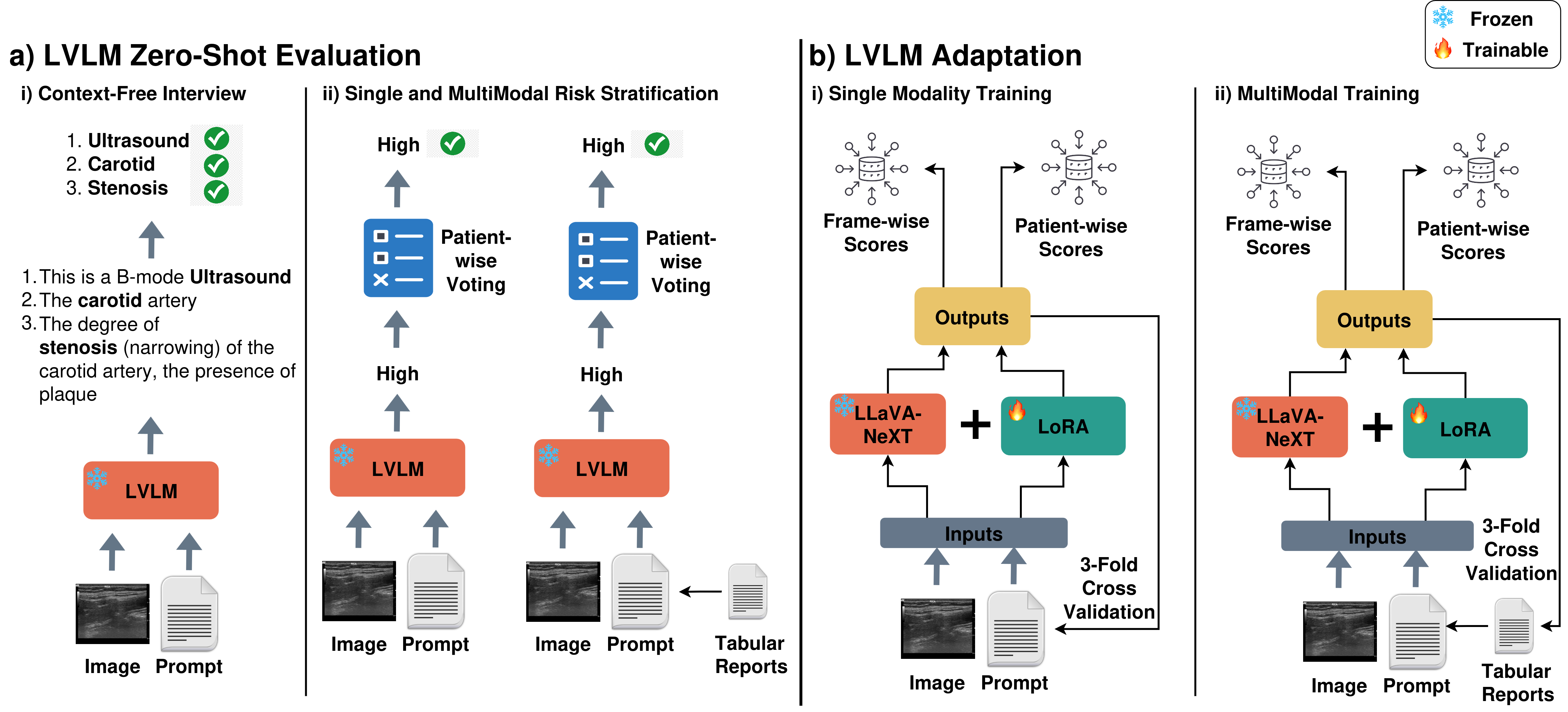}}
\caption{Overview of the methods that were applied in this study. 1) A zero-shot evaluation framework was designed to assess state-of-the-art large vision language models in i) medical reasoning and ii) risk stratification using single and multimodal inputs from carotid ultrasound images and clinical, demographic, laboratory, and protein analysis tabular data. 2) The adaptation of LLaVA-NeXT is performed using Low-rank Adaptation (LoRA) and 3-fold cross validation in a single and multimodal setting.}
\label{fig:bibe_flow}
\end{figure*}

In medicine, the ability of LVLMs to leverage multimodal information makes them particularly promising for complex diagnostic tasks where clinical decision-making depends on the synthesis of heterogeneous data types, such as imaging features, laboratory results, demographic factors, and clinical information \cite{shaaban2024medpromptx}. However, despite this potential, their application in this domain, particularly in the field of Ultrasound imaging (USI), remains largely unexplored and challenging \cite{llaus, u2bench}. USI is a particularly challenging modality, especially for AI-based systems. It is highly operator-dependent, susceptible to artifacts such as speckle noise and low contrast, and its appearance can vary considerably with acquisition settings, patient anatomy, and device manufacturer \cite{llaus, u2bench}. Furthermore, most LVLMs are pre-trained on natural images and everyday concepts \cite{llava-next, llava1, paligemma2}, making direct transfer to USI analysis not easy. The lack of large and high-quality paired image–text datasets in ultrasound, coupled with the fact that most existing multimodal medical datasets contain little or no USI data alongside rich clinical context, further hinders progress. This creates a pressing need to assess how well current LVLMs can adapt to the ultrasound domain and leverage associated textual data for clinically meaningful reasoning.

Several recent works have explored this direction. General-purpose open-source LVLMs such as PaliGemma \cite{paligemma2} and LLaVa \cite{llava1, llava2, llava3} have improved rapidly in recent years, in some cases approaching or even surpassing the performance of commercial models such as GPT-4V \cite{gpt4} and Gemini \cite{gemini} on certain multimodal benchmarks. Their main advantage lies in the ability to be fine-tuned or adapted to specific datasets and domains, enabling improved performance in medical tasks and fostering research progress. MedGemma \cite{medgemma}, developed by Google Deepmind, extends the Gemma 3 \cite{gemma3} language model with a medically tuned vision encoder, demonstrating strong multimodal medical reasoning capabilities across images and text, and serving as a solid foundation for downstream clinical applications. LLaVa-Med \cite{llavamed} is another example of a medical-domain LVLM, fine-tuned from LLaVA using biomedical figures, captions, and instruction-style data, and has been applied to build interactive medical assistants. In the histopathology domain, QUILT-LLaVA \cite{quiltllava} is a LLaVa-based model fine-tuned on pathology image–text pairs derived from expert training videos, incorporating positional annotations of regions of interest to improve spatial grounding. Although these approaches have shown promise in radiology, pathology and ophthalmology, progress in USI has been slower, with only a few recent attempts such as LLAUS \cite{llaus}, an instruction-tuned LVLM for obstetric and general ultrasound.

Benchmarking efforts also play a crucial role in this field. In general medical multimodal reasoning, GMAI-MMBench \cite{gmai-bench} serves as a large-scale benchmark covering diverse medical imaging modalities and specialties, designed to test both perception and high-level reasoning in LVLMs. However, it contains a relatively small number of ultrasound cases. To address this, U2-Bench \cite{u2bench} was recently introduced as a comprehensive USI benchmark that offers a multi-task evaluation framework, spanning 15 anatomies such as breast, heart and lung, and 8 clinical tasks like classification (i.e. disease diagnosis) and text generation. This benchmark enables a more rigorous assessment of model capabilities in this modality. Even if U2-Bench is very diverse, it does not really evaluate the challenging task of carotid risk stratification.

Cardiovascular diseases (CVD) represent a significant clinical challenge, with carotid atheromatous plaques being one of the leading causes of ischemic stroke worldwide \cite{bos2021atherosclerotic}. Atherosclerosis is a chronic and progressive inflammatory disease of the arterial wall, characterized by lipid accumulation, immune cell infiltration, and the formation of a fibrous cap over a lipid-rich core \cite{melaku2021cellular, malekmohammad2021role}. Over time, these plaques can become unstable due to several risk factors, increasing the risk of rupture and subsequent cerebrovascular events such as stroke. Accurate risk stratification of carotid plaques is crucial for determining appropriate therapeutic interventions and preventing cerebrovascular events. Traditional approaches to plaque assessment have relied primarily on single-modality imaging, typically USI \cite{ganitidis2021stratification} or computed tomography angiography \cite{antonopoulos2022cardiovascular}, combined with clinical risk factors \cite{li2025research}. However, the complex nature of atherosclerotic disease necessitates a more comprehensive approach that integrates diverse data sources to achieve optimal risk prediction, while ensuring that the underlying reasoning remains transparent and comprehensible to humans \cite{shi2023radiomics, liapi2024carotid, wang2022identification}.

This study investigates the capabilities of state-of-the-art (SOTA) LVLMs on detecting vulnerable atheromatous plaques from multimodal data. Open-source models, such as LLaVA-Next \cite{llava-next}, are employed, focusing on the integration of USI with free text-based information, including clinical, demographic, laboratory, and protein analysis data. To simulate realistic diagnostic scenarios, a set of tailored prompts is designed that incorporates both image and textual patient-specific clinical context. The presented evaluation framework begins with an interviewing prompt sequence, investigating each model's visual understanding of the USI modality, the organ displayed, and the reasoning behind patient-specific risk factors.  Subsequently, the models are further assessed in stroke risk stratification, following a zero-shot, as well as a task-specific fine-tuning approach.

An overview of the study can be seen in Fig. \ref{fig:bibe_flow}. In particular, the main contributions of this study are twofold: \begin{enumerate*} \item the first in-depth comparative evaluation of general-purpose and medically tuned LVLMs for medical reasoning and inference for carotid plaque risk stratification, and \item the task-specific fine-tuning of LLaVa-NeXT-Vicuna on carotid USI data, addressing the current scarcity of domain-adapted LVLMs for ultrasound-based CVD risk stratification \end{enumerate*}. Finally, the study provides empirical insights into the strengths and limitations of existing LVLMs in medical inference, highlighting key challenges for generalization to real-world clinical settings.

\section{Methods}
\subsection{Dataset and image strategy}
In this study, an anonymized private dataset 
was used, consisting of B-mode carotid USI sequences (videos) with corresponding tabular data from $72$ patients, aged 56 to 80 years, all presenting with $>30\%$ atherosclerotic plaque. All videos were acquired under standardized conditions (including patient preparation, examination position, room temperature, and transducer settings) with a minimum duration of $3$ \textit{seconds}. A frame rate exceeding $25$ \textit{frames per second} was used, capturing approximately $2$ to $3$ cardiac cycles. The image resolution was $12$ \textit{pixels per millimetre} in both the radial and longitudinal directions. Each video was labeled by the degree of stenosis and the presence of a stroke or transient ischaemic attack within the six months preceding the ultrasound examination. Each sample was classified as high risk if it met one of the following criteria: \begin{enumerate}
\item carotid artery stenosis exceeding $70\%$, or
\item occurrence of stroke or transient ischaemic attack symptoms within the preceding six months.
\end{enumerate} The dataset includes $59$ high-risk and $13$ low-risk patients, making it suitable for binary classification but inherently imbalanced. The tabular data included clinical and demographic information such as age, sex, and prescribed medications with dosages, blood test results, and protein analysis. None of the patients exhibited additional health conditions, such as cancer, chronic disease, or heart failure. 

This video-based dataset was pre-processed to generate image-text pairs suitable for LVLM input. To maximize the use of each patient’s video and to extract as much as possible visual information for zero-shot evaluation and training, multiple frames were extracted using two distinct sampling strategies:
\begin{enumerate}
    \item \textbf{Random Sampling.} Five frames were randomly selected from each patient’s video, after first dividing the video into five equal segments using a deterministic process. This ensured coverage across different phases of the cardiac cycle—during which the plaque might be more visible—and increased dataset variability at the frame level. These frames were then used for the zero-shot evaluation of the selected LVLMs.
    \item \textbf{Full-frame Sequence.} In the second sampling strategy, all frames from each video were used for model adaptation. Training and testing were performed on disjoint patient subsets to prevent data leakage. Since videos had varying numbers of frames, some patients were overrepresented; this was mitigated with patient-level 3-fold cross-validation, and results are reported as the average across folds.
\end{enumerate}

\subsection{Prompt strategies for LVLMs}
\begin{figure*}[htbp]
\centerline{\includegraphics[scale=0.25,keepaspectratio=true]{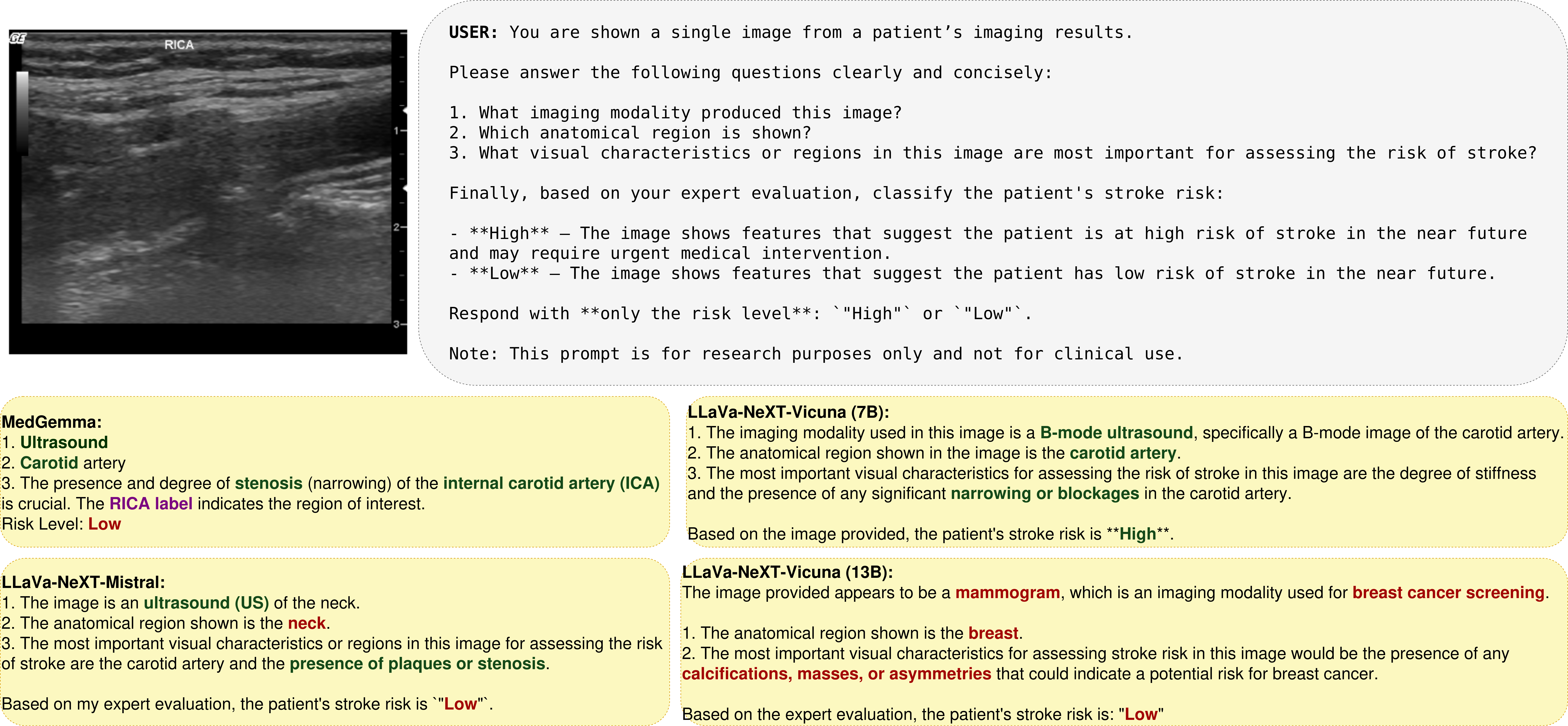}}
\caption{An example of the context-free interview process. The model is prompted with the image and the interview-style text. Each model produces different responses, indicating its zero-shot capabilities in medical reasoning.}
\label{fig:interview}
\end{figure*}

\begin{figure*}[htbp]
\centerline{\includegraphics[scale=0.25,keepaspectratio=true]{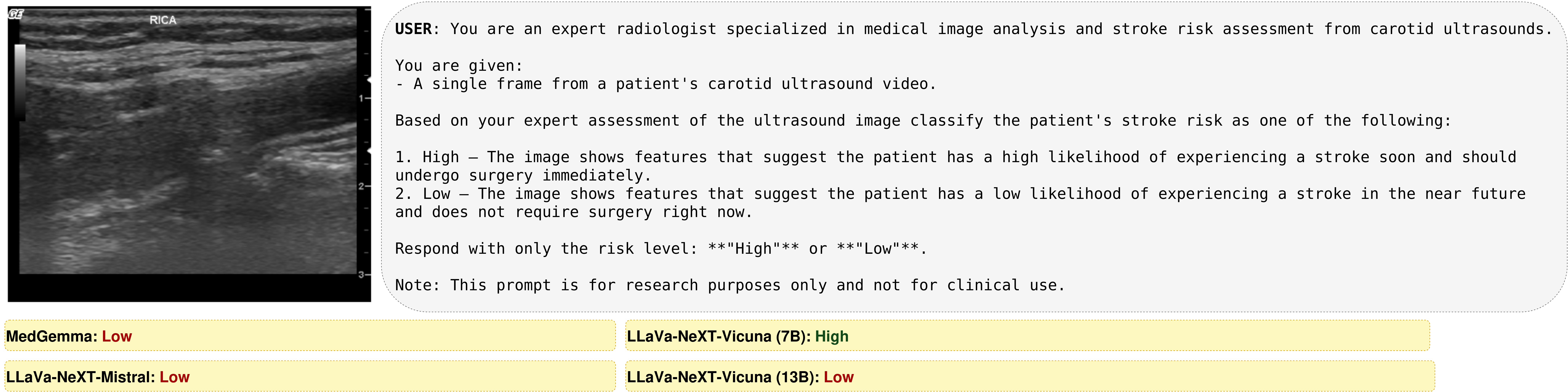}}
\caption{An example of the imaging context (single modal) prompt. The model is prompted with the image and user text that provides context for the image and asks a question that should be answered with a single word. This way of prompting is designed to assess the model's diagnosis capabilities from a single image.}
\label{fig:tempB}
\end{figure*}

\begin{figure*}[htbp]
\centerline{\includegraphics[scale=0.25,keepaspectratio=true]{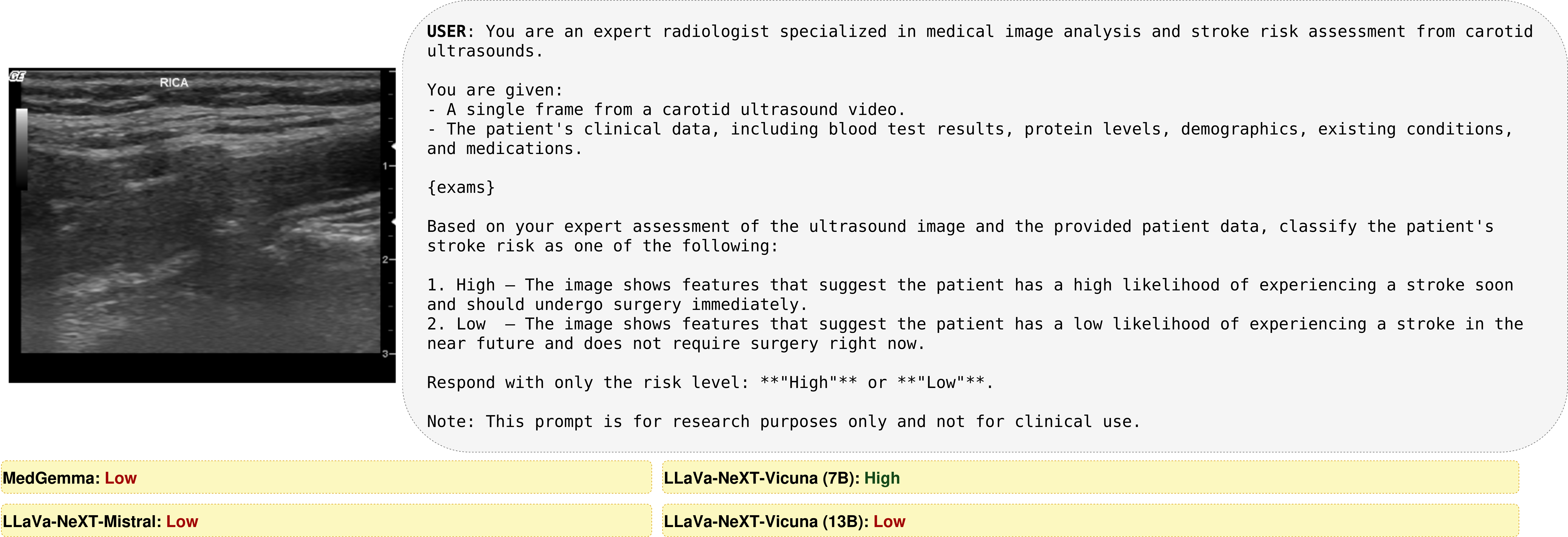}}
\caption{An example of the imaging and tabular (multimodal) context prompt. The model is prompted with the image and user text that provides context for the image, demographic, clinical, and laboratory information, inside the test and asks a question that should be answered with a single word. This way of prompting is designed to assess the model's diagnosis capabilities from a multimodal prompt.}
\label{fig:tempA}
\end{figure*}

 Three text prompts were designed to explore and compare the LVLMs’ capabilities in both zero-shot and fine-tuning settings:
 \begin{enumerate}
 \item \textbf{Context-Free Interview}: this prompt provided only the medical image and instructed the model to act as an expert radiologist (Fig. \ref{fig:interview}). The prompt contained COT-style questions about the modality that produced the input image, the depicted organ, and diagnostically relevant visual features.
     \item \textbf{Imaging Context}: This prompt included context about the imaging modality and organ (Fig: \ref{fig:tempB}). It also defined high and low risk classes using the criteria we described before and requested a classification into one of these classes.
     \item \textbf{Image and Tabular Context}: This prompt extended the image-only prompt, with patient specific demographic, clinical, and laboratory information converted to free text (Fig. \ref{fig:tempA}). Binary values replaced with descriptive terms, e.g. 0 and 1 turned to male and female, and blood test acronyms expanded to their full names.
 \end{enumerate}

All image–text pairs were tokenized and normalized according to each VLM’s input requirements using the relevant Hugging Face \cite{hf-trans} pre-processing modules.

\subsection{Benchmarking of the LVLMs models}



Our study first focuses on the investigation of the capabilities of LVLMs for ultrasound image understanding. Each model was prompted with the frames selected using the random sampling strategy for each patient, accompanied by the interview text. Three tasks were evaluated:
\begin{enumerate*}
    \item modality identification,
    \item anatomy identification, and
    \item risk stratification.
\end{enumerate*}

Models that achieved the highest accuracy in modality and anatomy identification without additional context were further tested on the risk stratification task using two prompt settings: single-modality setting using only image data and multimodal setting using image combined with tabular data. This aimed to investigate whether providing contextual information related to demographics, clinical risk factors, and laboratory examinations could improve decision-making in risk stratification. 

In this study, six open-source LVLMs were explored. Five were general-purpose models—trained primarily on natural image–text pairs and suitable for inference with free-text prompts—namely: \textit{Paligemma-Mix (3B) \cite{paligemma1}, Paligemma2-Mix (3B) \cite{paligemma2}, LLaVa-NeXT Vicuna (7B and 13B), LLaVa-NeXT Mistral (7B) \cite{llava-next}}. These models were selected for their strong performance on vision–language benchmarks, approaching that of leading commercial models such as \textit{GPT-4} \cite{gpt4}.
In addition, one medically tuned vision–language foundation model, \textit{MedGemma (4B)} \cite{medgemma} was included. This model was chosen for its improved performance on a range of medical imaging benchmarks \cite{medgemmacard, medgemmablog} and because, according to its documented training data composition \cite{medgemmacard}, it had not been trained on any ultrasound datasets, allowing for an unbiased assessment in this domain.

\begin{table*}[t!]
\caption{Prediction Rate of Most Mentioned Imaging Modalities and Anatomies for $4$ out of the $6$ LVLMs that were used in this study. The Paligemma family of models had failed to identify both the modality and anatomy of the input and for this reason were excluded from the table.}
\begin{center}
\begin{tabular}{l|l|l|l|l||l|l|l|l}
\hline
\multicolumn{1}{l|}{\multirow{2}{*}{\textbf{Models}}} & \multicolumn{4}{c||}{\textbf{Imaging Modalities}}                                                                   & \multicolumn{4}{c}{\textbf{Anatomies}}                                                                                    \\ \cline{2-9} 
\multicolumn{1}{l|}{}                        & \multicolumn{1}{l|}{\textbf{MRI}} & \multicolumn{1}{l|}{\textbf{X-Ray}} & \multicolumn{1}{l|}{\textbf{CT}} & \multicolumn{1}{l||}{\textbf{USI}} & \multicolumn{1}{l|}{\textbf{Brain}} & \multicolumn{1}{l|}{\textbf{Lung}} & \multicolumn{1}{l|}{\textbf{Neck}} & \multicolumn{1}{l}{\textbf{Carotid}} \\ \hline \hline
\textbf{LLaVa-Next-Vicuna-7B \cite{llava-next}}  & $0.56\%$& $0\%$& $3.89\%$& $\textbf{95\%}$   & $2.22\%$& $0.28\%$& $0.28\%$& $\mathbf{91.94\%}$ \\ \hline
\textbf{LLaVa-Next-Vicuna-13B \cite{llava-next}} & $0\%$& $96.94\%$& $3.06\%$ & $0\%$  &  - & - & - & - \\ \hline
\textbf{LLaVa-Next-Mistral \cite{llava-next}}   & $2.78\%$& $2.22\%$& $0\%$& $\mathbf{95\%}$   & $2.78\%$& $2.22\%$& $94.16\%$& $0.83\%$ \\ \hline
\textbf{MedGemma \cite{medgemma}}    & $0\%$ & $0\%$ & $0\%$& $\mathbf{100\%}$  & $0.28\%$ & $0\%$ & $0.56\%$&  $\mathbf{98.33\%}$ \\
\hline
\end{tabular}
\label{tab:modality_organ_pred}
\end{center}
\end{table*}
\subsection{LVLM Adaptation to USI data and Carotid Risk Stratification}
Following the zero-shot evaluation, the best-performing general-purpose model, \textit{LLaVa-NeXT Vicuna (7B)}, was selected and adapted for risk stratification using LoRA adapter \cite{lora}. This model was selected such that we will test the challenges and opportunities of general-purpose models. Moreover, LoRA adaptation is well-suited for large transformer-based models as it freezes pre-trained weights, injects small low-rank matrices into selected layers, and trains only these matrices. This approach is more memory- and time-efficient than full fine-tuning.

The LLaVa-Next Vicuna (7B) model was adapted in two scenarios: \begin{enumerate*}
    \item single-modality risk stratification using only image data as input and, \item multimodal risk stratification combining image and tabular data as input.
\end{enumerate*} In both cases, the full frame sequences of each patient were used. The aim was to investigate whether a LVLM can leverage multimodal datasets that combine images with structured patient information.

Training was performed using the official \texttt{LLaVaTrainer} module (available on GitHub\footnote{https://github.com/haotian-liu/LLaVA}) with LoRA rank $r = 128$ and alpha $\alpha = 128$. Since LLaVa is transformer-based, low-rank matrices were inserted into the attention projection matrices of the query and value components ($W_Q, W_V$) in both the vision encoder and the language model. Images were resized and padded to $336 \times 336$ pixels using the default LLaVa preprocessing pipeline.

The AdamW \cite{adamw} optimizer was used with a learning rate of $1\mathrm{e}{-5}$, a warm-up ratio of $3\%$, and a cosine learning rate scheduler. In LLaVA models, a Multilayer Perceptron (MLP) cross-modal connector aligns image and text features; this module was trained with a higher learning rate of $2\mathrm{e}{-4}$ to enable faster adaptation to the USI domain. Training employed the standard cross-entropy loss over output tokens, with all other hyperparameters kept as in the official LLaVA fine-tuning script. Model selection followed a patient-level 3-fold cross-validation scheme.

\subsection{Evaluation Protocol}
For zero-shot evaluation, interview responses were parsed using regular expressions to extract relevant information, and the proportion of cases in which the models explicitly mentioned the terms "ultrasound" and "carotid" was computed. For binary classification, Sensitivity, Specificity, Balanced Accuracy as well as Mathews Correlation Coefficient (MCC), and Area Under the Curve (AUC), known for their suitability in imbalanced data sets, were used as evaluation metrics.

These metrics were computed both \textit{frame-wise} and \textit{patient-wise}. For patient-wise metrics, majority voting over frame-level predictions was applied. AUC was calculated only at the patient level. In order to obtain a probability prediction, the mean value across \textit{frame-wise} label predictions was calculated and used for the AUC evaluation.




\begin{table*}[t!]
\caption{Average evaluation scores for 3-fold cross validation after the adaptation of the  LLaVa-NeXT-Vicuna-7B  model. The evaluation was performed both in frame level and patient level after averaging the probabilities received for the frame level. The evaluation was performed in terms of Sensitivity (Sens.), Specificity (Spec.), Balanced
Accuracy (bAcc) as well as Mathews Correlation Coefficient (MCC),
and Area Under the Curve (AUC).}
\begin{center}
\scalebox{0.8}{
\begin{tabular}{l|l|l|l|l||l|l|l|l|l}
\hline
\multicolumn{1}{l|}{\multirow{2}{*}{\textbf{Tasks}}} & \multicolumn{4}{c||}{\textbf{Frame-Level Evaluation}}                                                                   & \multicolumn{5}{c}{\textbf{Patient-Level Evaluation}}                                                                                    \\ \cline{2-10} 
\multicolumn{1}{l|}{}                        & \multicolumn{1}{l|}{\textbf{bAcc}} & \multicolumn{1}{l|}{\textbf{Sens.}} & \multicolumn{1}{l|}{\textbf{Spec.}} & \multicolumn{1}{l||}{\textbf{MCC}} & \multicolumn{1}{l|}{\textbf{bAcc}} & \multicolumn{1}{l|}{\textbf{Sens.}} & \multicolumn{1}{l|}{\textbf{Spec.}} & \multicolumn{1}{l|}{\textbf{MCC}}  &
\multicolumn{1}{l}{\textbf{AUC}} 
\\ \hline \hline
Single-modality (Image)& $71.3\pm11.8\%$& $94.2\pm2.6\%$& $48.4\pm25\%$& $42.9 \pm 16.4\%$ & $73.2\pm8.1\%$& $91.5\pm3.1\%$& $55\pm18\%$& $47.4 \pm 1.1\%$ & $82.6 \pm 11\%$ 
\\
\hline
Multimodal (Image \& Tabular)& $74.5\pm17\%$& $94.7\pm 6.1\%$& $54.3\pm40\%$& $50\pm12\%$
& $77.5\pm13.3\%$& $91.6\pm5\%$& $63.3\pm32.1\%$& $54.4\pm13.8\%$ & $\mathbf{84.3\pm13.5}$\\
\hline
CNN-Ensemble \cite{ganitidis2021stratification} & -- & -- & -- & -- & $72.5\pm6\%$& $75\pm17.6\%$ & $70\pm10.3\%$& -- &$73\pm 10.7\%$\\
\hline
\end{tabular}
\label{tab:frame-patient-cls-scores}}
\end{center}
\end{table*}

\section{Results}
\subsection{Zero-Shot Evaluation}
\subsubsection{Imaging Modality Detection}
Table ~\ref{tab:modality_organ_pred} presents the frequency with which different medical imaging modalities were mentioned by the models during the zero-shot interview evaluation. The medically tuned model, MedGemma, correctly identified USI in all cases. Among the general-purpose models, the smaller LLaVa variants (LLaVa-NeXT-Vicuna-7B and LLaVa-NeXT-Mistral) also achieved high accuracy, correctly identifying USI in $95\%$ of cases; in the remaining 5\%, they predicted other modalities such as magnetic resonance imaging (MRI), X-ray, or computed tomography (CT). In contrast, the larger LLaVa-NeXT-Vicuna-13B model consistently failed to detect the correct modality.

The Paligemma family of models is not included in this table, as both models consistently returned the message: \textit{"Sorry, as a base VLM I am not trained to answer this question."}—even when a disclaimer was included in the prompt specifying that responses were for research purposes and not medical advice. Further zero-shot testing revealed that these models were highly sensitive to prompt wording, with minor changes in phrasing or syntax dramatically altering their outputs. When asked solely to describe the image content, they correctly identified the modality in roughly $35\%$ of the cases. This highlights that Paligemma models are not suitable for USI tasks in a zero-shot setting.

\subsubsection{Anatomy Detection}
Table ~\ref{tab:modality_organ_pred} shows the frequency of some of the most common anatomies mentioned by the models that correctly predicted the USI modality. The two LLaVa variants demonstrated completely different patterns in their answers. LLaVa-NeXT-Vicuna-7B correctly mentioning the carotid artery $91.94\%$ of cases, whereas LLaVa-NeXT-Mistral did so in only $0.83\%$ of cases,  instead referring to the neck in $94.16\%$ of cases. Overall, LLaVa-NeXT-Vicuna-7B was the most accurate general-purpose model in this study. MedGemma again outperformed all others, identifying the correct anatomy in $98.33\%$ of cases. 

A closer qualitative inspection of the best-performing models revealed the specific visual cues that informed their predictions. When the carotid artery was correctly identified, the models often referenced clinical keywords such as “stenosis”, “narrowing”, and “plaque”. Notably, both models sometimes leveraged textual labels embedded in the ultrasound images themselves. In particular, the label “ICA” (and its variants “LICA” and “RICA”), originating from the raw videos and denoting the Left/Right Internal Carotid Artery, was occasionally detected and used in reasoning.

However, the models differed in how they interpreted this label. LLaVa-NeXT-Vicuna-7B detected “LICA” in only $7$ out of $360$ images, and incorrectly expanded it as Lateral Intra-Cranial Artery. In contrast, MedGemma frequently identified the ICA/LICA/RICA labels correctly and appeared to base almost all of its predictions primarily on this textual cue rather than on the intrinsic visual features of the carotid artery. This reliance suggests a potential label-induced bias, meaning the model’s strong performance in most cases may have been aided by incidental on-image text rather than true anatomical recognition.



\subsection{Zero-shot Classification}
All models failed the zero-shot risk stratification task. LLaVa-NeXT-Vicuna-7B consistently predicted \textit{High risk} for every case, while LLaVa-NeXT-Mistral predicted \textit{Low risk} across the board. Even MedGemma exhibited a similar bias, predicting \textit{Low risk} in $99.44\%$ of cases. The Paligemma models were prompted with slightly different instructions for the same tasks to encourage them to respond with the class label; however, their classification performance was worse than that of a random classifier. This finding validates that USI data are not easy to be handled in a zero-shot fashion from LVLMs and their adaptation is needed for more specific tasks.

\subsection{Model Adaptation}
Table ~\ref{tab:frame-patient-cls-scores} presents the average evaluation metrics and standard deviation from the 3-fold cross-validation of LLaVa-NeXT-Vicuna-7B on the held-out patient cases, for both frame- and patient-level evaluation, under the \text{single-modality} and \textit{multimodal} settings. The results indicate that incorporating tabular metadata into the multimodal framework improves risk stratification performance. In particular, at the patient-level (Table ~\ref{tab:frame-patient-cls-scores}, right), the specificity of the \textit{multimodal} setting ($63.3 \pm 32.1\%$) exceeds that of the \textit{single-modality} setting ($55 \pm 18\%$), suggesting that tabular data help the model better identify the under-represented class. Overall, these findings show that the adapted model can effectively leverage the image-text pairs provided during training.



An interesting result is that, across both frame-wise and patient-wise evaluations, specificity values demonstrate large standard deviations, indicating high-variability in low-risk detection across folds. This is likely the result of the strong class imbalance present in the dataset. In contrast, the sensitivity values remained consistently high ($>90\%$) with low standard deviations. This reveals a  strong bias toward predicting \textit{High risk} class.

For context, Table ~\ref{tab:frame-patient-cls-scores} also includes results from prior SOTA work \cite{ganitidis2021stratification} on the same task, which trained a set of CNN models from scratch on the same dataset using radiologist-segmented cropped images, and used them in an ensemble setting for inference. That study employed 4-fold cross-validation and included additional patient cases (patients with multiple videos), which were excluded in the current experiments to reduce bias. While the differences in dataset composition and validation setup prevent direct comparison, these prior results provide a reference for the performance of an expert, domain-specific model on the same task.

Comparing the reported scores, the adapted LLaVa-NeXT-Vicuna-7B achieves a higher AUC to the CNN ensemble but exhibits lower specificity, indicating greater difficulty in detecting the low-risk class. Combined with the relatively large standard deviations observed across folds and the high sensitivity scores, these findings suggest that the adapted model’s generalizability remains limited.

\section{Discussion}
This work provides a systematic evaluation of state-of-the-art open-source LVLMs on ultrasound-based cardiovascular risk stratification, integrating USI with structured clinical, demographic, and laboratory data in textual form.

Our findings in the zero-shot setting, show that while current LVLMs can identify imaging modality and anatomical region with high accuracy in most cases, their performance on clinically meaningful classification tasks remains poor. MedGemma and LLaVa-NeXT-Vicuna-7B demonstrated strong performance in ultrasound modality detection and carotid artery recognition. Qualitative inspection revealed that when the models correctly recognized the carotid artery, they related it with common clinical keywords. However, for the MedGemma model, part of this performance can be attributed to textual cues such as embedded textual labels in the ultrasound images. This label-induced bias raises concerns about overestimating true visual reasoning ability when models rely on such cues, rather than visual features. Importantly, all evaluated LVLMs failed in the zero-shot classification of plaque vulnerability, predicting almost the entire test set to a single class. This might reflect the absence of domain-specific priors in pretraining and the difficulty to relate ultrasound features to clinical risk factors without explicit supervision.

Adapting LLaVa-NeXT-Vicuna-7B via LoRA fine-tuning significantly improved risk stratification performance. The integration of tabular data consistently increased specificity and balanced accuracy compared to the image single-modality setting, suggesting that LVLMs can benefit from multimodal tabular data when available. Nevertheless, specificity exhibited high variance across folds, suggesting unstable low-risk detection likely driven by severe class imbalance. In contrast, sensitivity remained above $90\%$ in all cases, with low variance, indicating strong high-risk class detection. This also suggests a tendency towards the high-risk class, caused by the class imbalance.

When compared to prior CNN-based models trained on cropped images from the same dataset, the fine-tuned LVLM achieved comparable or higher AUC but lower specificity. This suggests that the fine-tuned LVLM matches overall discriminative power, but remains less calibrated for label prediction. However, the ensemble’s reliance on expert-annotated data poses a significant requirement for manual effort, domain expertise, and time-consuming segmentation workflows, limiting its scalability and applicability in real-world clinical environments where such annotations are rarely available.

This study shows that, despite the challenges pointed above, a fine-tuned LVLM can achieve competitive discriminative performance, even matching or exceeding the AUC of CNN baselines trained on expert-annotated data. The model’s ability to leverage both imaging and structured clinical data demonstrates the strength of multimodal integration, with significant performance improvement when patient metadata are included. These results highlight the adaptability of general purpose LVLMs to medical domains with minimal domain-specific training, offering a scalable path toward flexible diagnostic systems.

Potential limitations of the proposed work include the dataset size and class imbalance, both of which, while realistic, impact the stability and generalizability of results. Additionally, processing a video-based dataset as a collection of static images overlooks temporal information that is often crucial for diagnosis, as medical experts frequently rely on motion cues when performing risk stratification. Another limitation lies in the heterogeneity of the input modalities, integrating ultrasound frames with tabular patient data poses challenges for model alignment, and the current LVLM architectures may not optimally handle such multimodal fusion without architectural modifications. Moreover, the benchmark evaluation may be influenced by prompt design, and differences in how models handle medical terminology, potentially affecting comparability. Finally, fine-tuning on a small, domain-specific dataset risks overfitting, especially given the high model capacity of LVLMs.

Building on the results of this study, several key areas should be addressed towards translating LVLMs into reliable clinical tools. First, expanding to larger, and more balanced multimodal datasets will be crucial to improve model robustness and mitigate biases originated from limited training data. Improving calibration, particularly for underrepresented classes, will enhance the reliability of predictions in real-world decision-making. Incorporating temporal features from ultrasound sequences could further enrich the models’ understanding of vascular motion through the cardiac cycle and improve diagnostic accuracy. Finally, future directions should leverage the inherent interactive capabilities of LVLMs, such as question–answering sequences, to provide step-by-step reasoning. Such features can enhance interpretability, clarify diagnostic decisions, and support the integration of these models into real-world diagnostic workflows.

\section{Conclusion}
This study presents an in-depth evaluation of open-source LVLMs for carotid plaque risk stratification from ultrasound images combined with structured clinical data. Zero-shot evaluation revealed that while some models excel at identifying imaging modality and anatomy, they fail at clinically relevant classification without adaptation. Fine-tuning LLaVa-NeXT-Vicuna-7B using LoRA significantly improved performance, especially when multimodal tabular data were included, achieving competitive AUC against CNN baseline architectures.

\section*{Acknowledgment}
The authors acknowledge support from GRNET – National Infrastructures for Research and Technology, which provided access to AWS cloud infrastructure. This work has been partially supported by project MIS 5154714 of the National Recovery and Resilience Plan Greece 2.0, funded by the European Union under the NextGenerationEU Program, administered by the Archimedes Unit of the Athena Research Center.






\vspace{12pt}


\begin{thebibliography}{00}

\bibitem{li2025vision} X. Li et al., “Vision-Language Models in medical image analysis: From simple fusion to general large models,” Information Fusion, p. 102995, 2025.

\bibitem{vit} A. Dosovitskiy et al., “An image is worth 16x16 words: Transformers for image recognition at scale,” arXiv preprint arXiv:2010.11929, 2020.

\bibitem{llava-next} H. Liu et al., “LLaVA-NeXT: Improved reasoning, OCR, and world knowledge.” Jan. 2024. [Online]. Available: https://llava-vl.github.io/blog/2024-01-30-llava-next/

\bibitem{wei2023cot} J. Wei et al., “Chain-of-Thought Prompting Elicits Reasoning in Large Language Models,” Jan. 10, 2023, arXiv: arXiv:2201.11903. doi: 10.48550/arXiv.2201.11903.

\bibitem{xu2025llavacot} G. Xu et al., “LLaVA-CoT: Let Vision Language Models Reason Step-by-Step,” July 21, 2025, arXiv: arXiv:2411.10440. doi: 10.48550/arXiv.2411.10440.

\bibitem{shaaban2024medpromptx} M. A. Shaaban, A. Khan, and M. Yaqub, ``MedPromptX: Grounded Multimodal Prompting for Chest X-Ray Diagnosis,'' in Medical Image Computing and Computer Assisted Intervention – MICCAI 2024 Workshops, A. Schroder, X. Li, T. Syeda-Mahmood, N. P. Oxtoby, A. Young, A. Hering, T. S. Mathai, P. Mukherjee, S. Kuckertz, T. He, I. Llorente-Saguer, A. Maier, S. Kashyap, H. Greenspan, and A. Madabhushi, Eds., Cham: Springer Nature Switzerland, 2025, pp. 211–222. doi: 10.1007/978-3-031-84525-3\_18.
\bibitem{llaus} J. Guo, X. Shan, G. Wang, D. Chen, R. Lu, and S. Tang, “LLAUS: A High-Quality Instruction-Tuned Large Vision Language Assistant for UltraSound,” in Proceedings of the 2025 International Conference on Multimedia Retrieval, Chicago IL USA: ACM, June 2025, pp. 398–406. doi: 10.1145/3731715.3733374.

\bibitem{u2bench} A. Le et al., “U2-BENCH: Benchmarking Large Vision-Language Models on Ultrasound Understanding,” arXiv preprint arXiv:2505.17779, 2025, unpublished.

\bibitem{llava1} A. Radford et al., “Learning Transferable Visual Models From Natural Language Supervision,” Feb. 26, 2021, arXiv: arXiv:2103.00020. doi: 10.48550/arXiv.2103.00020.

\bibitem{paligemma2} A. Steiner et al., “Paligemma 2: A family of versatile vlms for transfer,” arXiv preprint arXiv:2412.03555, 2024.

\bibitem{bos2021atherosclerotic} D. Bos et al., “Atherosclerotic Carotid Plaque Composition and Incident Stroke and Coronary Events,” JACC, vol. 77, no. 11, pp. 1426–1435, Mar. 2021, doi: 10.1016/j.jacc.2021.01.038.

\bibitem{ganitidis2021stratification} T. Ganitidis, M. Athanasiou, K. Dalakleidi, N. Melanitis, S. Golemati, and K. S. Nikita, “Stratification of carotid atheromatous plaque using interpretable deep learning methods on B-mode ultrasound images,” in 2021 43rd Annual International Conference of the IEEE Engineering in Medicine \& Biology Society (EMBC), 2021, pp. 3902–3905. doi: 10.1109/EMBC46164.2021.9630402.

\bibitem{antonopoulos2022cardiovascular} A. S. Antonopoulos, A. Angelopoulos, K. Tsioufis, C. Antoniades, and D. Tousoulis, “Cardiovascular risk stratification by coronary computed tomography angiography imaging: current state-of-the-art,” Eur. J. Prev. Cardiol., vol. 29, no. 4, pp. 608–624, Mar. 2022, doi: 10.1093/eurjpc/zwab067.

\bibitem{li2025research} Z.-L. Li et al., “Research on ischemic stroke risk assessment based on CTA radiomics and machine learning,” BMC Med Imaging, vol. 25, no. 1, p. 206, Jun. 2025, doi: 10.1186/s12880-025-01697-y.


\bibitem{llava2} H. Liu, C. Li, Q. Wu, and Y. J. Lee, “Visual Instruction Tuning,” Dec. 11, 2023, arXiv: arXiv:2304.08485. doi: 10.48550/arXiv.2304.08485.

\bibitem{llava3} H. Liu, C. Li, Y. Li, and Y. J. Lee, “Improved Baselines with Visual Instruction Tuning,” May 15, 2024, arXiv: arXiv:2310.03744. doi: 10.48550/arXiv.2310.03744.

\bibitem{gpt4} OpenAI et al., “GPT-4 Technical Report,” Mar. 04, 2024, arXiv: arXiv:2303.08774. doi: 10.48550/arXiv.2303.08774.

\bibitem{gemini} G. Team et al., “Gemini: A Family of Highly Capable Multimodal Models,” May 09, 2025, arXiv: arXiv:2312.11805. doi: 10.48550/arXiv.2312.11805.

\bibitem{medgemma} A. Sellergren et al., “Medgemma technical report,” arXiv preprint arXiv:2507.05201, 2025.

\bibitem{gemma3} G. Team et al., “Gemma 3 Technical Report,” Mar. 25, 2025, arXiv: arXiv:2503.19786. doi: 10.48550/arXiv.2503.19786.

\bibitem{llavamed} C. Li et al., “LLaVA-Med: Training a Large Language-and-Vision Assistant for Biomedicine in One Day,” June 01, 2023, arXiv: arXiv:2306.00890. doi: 10.48550/arXiv.2306.00890.

\bibitem{quiltllava} M. S. Seyfioglu, W. O. Ikezogwo, F. Ghezloo, R. Krishna, and L. Shapiro, “Quilt-LLaVA: Visual Instruction Tuning by Extracting Localized Narratives from Open-Source Histopathology Videos,” Jan. 13, 2025, arXiv: arXiv:2312.04746. doi: 10.48550/arXiv.2312.04746.

\bibitem{gmai-bench} J. Ye et al., “Gmai-mmbench: A comprehensive multimodal evaluation benchmark towards general medical ai,” Advances in Neural Information Processing Systems, vol. 37, pp. 94327–94427, 2024.

\bibitem{melaku2021cellular} L. Melaku and A. Dabi, “The cellular biology of atherosclerosis with atherosclerotic lesion classification and biomarkers,” Bulletin of the National Research Centre, vol. 45, no. 1, p. 225, Dec. 2021, doi: 10.1186/s42269-021-00685-w.

\bibitem{malekmohammad2021role} K. Malekmohammad, E. E. Bezsonov, and M. Rafieian-Kopaei, “Role of Lipid Accumulation and Inflammation in Atherosclerosis: Focus on Molecular and Cellular Mechanisms,” Front. Cardiovasc. Med., vol. 8, Sep. 2021, doi: 10.3389/fcvm.2021.707529.

\bibitem{shi2023radiomics} J. Shi et al., “Radiomics Signatures of Carotid Plaque on Computed Tomography Angiography,” Clin Neuroradiol, vol. 33, no. 4, pp. 931–941, Dec. 2023, doi: 10.1007/s00062-023-01289-9.

\bibitem{liapi2024carotid} G. D. Liapi et al., “Carotid Plaque Motion Analysis in Ultrasound Videos to Discover Rupture-Prone Plaque Areas,” in 2024 IEEE International Symposium on Biomedical Imaging (ISBI), Feb. 2024, pp. 1–4. doi: 10.1109/ISBI56570.2024.10635865.

\bibitem{wang2022identification} Y. Wang, T. Wang, Y. Luo, and L. Jiao, “Identification Markers of Carotid Vulnerable Plaques: An Update,” Biomolecules, vol. 12, no. 9, p. 1192, Sep. 2022, doi: 10.3390/biom12091192.

\bibitem{hf-trans} T. Wolf et al., “HuggingFace’s Transformers: State-of-the-art Natural Language Processing,” July 14, 2020, arXiv: arXiv:1910.03771. doi: 10.48550/arXiv.1910.03771.

\bibitem{paligemma1} L. Beyer et al., “Paligemma: A versatile 3b vlm for transfer,” arXiv preprint arXiv:2407.07726, 2024.

\bibitem{medgemmacard} “MedGemma model card | Health AI Developer Foundations,” Google for Developers. Accessed: Aug. 15, 2025. [Online]. Available: https://developers.google.com/health-ai-developer-foundations/medgemma/model-card

\bibitem{medgemmablog} “MedGemma: Our most capable open models for health AI development.” Accessed: Aug. 15, 2025. [Online]. Available: https://research.google/blog/medgemma-our-most-capable-open-models-for-health-ai-development/


\bibitem{lora} E. J. Hu et al., “Lora: Low-rank adaptation of large language models.,” ICLR, vol. 1, no. 2, p. 3, 2022.

\bibitem{adamw} I. Loshchilov and F. Hutter, “Decoupled Weight Decay Regularization,” Jan. 04, 2019, arXiv: arXiv:1711.05101. doi: 10.48550/arXiv.1711.05101.



\end{thebibliography}
\end{document}